\documentclass[a4paper,conference]{IEEEtran}
\ifCLASSINFOpdf
  \usepackage[pdftex]{graphicx}
\else
\fi
\usepackage{amsmath}
\usepackage{amssymb}
\usepackage{float}

\usepackage{gensymb}

\usepackage{booktabs}
\begin{document}
%
\title{Norm Loss: An efficient yet effective regularization method for deep neural networks}

\author{\IEEEauthorblockN{Theodoros Georgiou\IEEEauthorrefmark{1},
Sebastian Schmitt\IEEEauthorrefmark{2},
Thomas B\"ack\IEEEauthorrefmark{1},
Wei Chen\IEEEauthorrefmark{1} and
Michael Lew\IEEEauthorrefmark{1}}
\IEEEauthorblockA{\IEEEauthorrefmark{1}Leiden University\\
Leiden Institute of Advanced Computer Science,
Leiden, the Netherlands\\ Email: t.k.georgiou@liacs.leidenuniv.nl}
\IEEEauthorblockA{\IEEEauthorrefmark{2}Honda Research Institute Europe GmbH\\
Offenbach, Germany}}


%


\maketitle

\begin{abstract}
Convolutional neural network training can suffer from diverse issues like exploding or vanishing gradients, scaling-based weight space symmetry and covariant-shift. In order to address these issues, researchers develop weight regularization methods and activation normalization methods. In this work we propose a weight soft-regularization method based on the Oblique manifold. The proposed method uses a loss function which pushes each weight vector to have a norm close to one, i.e. the weight matrix is smoothly steered toward the so-called Oblique manifold. We evaluate our method on the very popular CIFAR-10, CIFAR-100 and ImageNet 2012 datasets using two state-of-the-art architectures, namely the ResNet and wide-ResNet. Our method introduces negligible computational overhead and the results show that it is competitive to the state-of-the-art and in some cases superior to it. Additionally, the results are less sensitive to  hyperparameter settings such as batch size and regularization factor.
\end{abstract}


%
\IEEEpeerreviewmaketitle

\section{Introduction}

Convolutional neural networks, and deep learning in general, have received a lot of attention in the past few years \cite{krizhevsky2012imagenet,he2016deep,BMVC2016_87,guo2018review,georgiou2019survey} and have been applied in many research areas, such as image understanding \cite{guo2018review,georgiou2019survey}, natural language processing \cite{young2018recent,ruder2019transfer} and game solving \cite{mnih2015human}. The research in these models has been motivated by the success of deep learning in image classification with AlexNet \cite{krizhevsky2012imagenet} and later by much deeper models such as VGG \cite{DBLP:journals/corr/SimonyanZ14a}, ResNet \cite{he2016deep} and wide-ResNet \cite{BMVC2016_87}. In order to understand these models and enhance their performance, a lot of research has been carried out and in many directions \cite{guo2018review,georgiou2019survey}, including data preprocessing strategies \cite{cubuk2019autoaugment,lim2019fast}, activation normalization \cite{wang2018kalman,wu2018group}, weight regularization \cite{bansal2018can,huang2020controllable}, activation functions \cite{trottier2017parametric}, and overall network architectures \cite{he2016deep,BMVC2016_87,han2017deep}. In this work we focus on weight regularization methods and propose a new soft regularization loss function, the norm loss.\par
Weight regularization has been a wide research topic. The main issues of deep learning training procedures are the exploding/vanishing gradients, as well as the scaling-based weight space symmetry and covariant-shift \cite{pmlr-v37-ioffe15,huang2017projection,bansal2018can}. In order to alleviate one or more of the aforementioned issues, researchers are developing methods that restrict the search space of possible weight vectors.\par
\begin{figure}[t!]
\centering
\includegraphics[width=.6\linewidth]{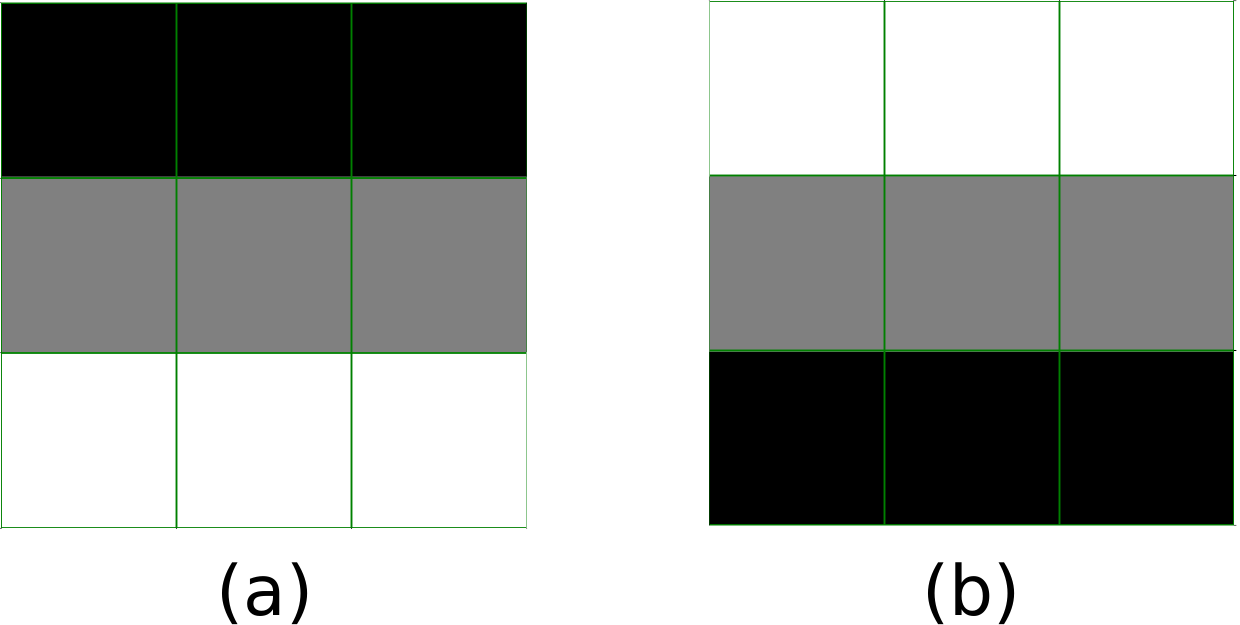}
\caption{ Schematic visualization of the weights of two example $3\times3$ one-channel kernels. White (dark) values indicate high (low) numbers. The weight matrix in example (b) is a 180\degree rotated version of (a).}
\label{fig:exKernels}
\end{figure}

One of the first approaches, and most popular, is the weight decay \cite{krizhevsky2012imagenet}, which adds a penalty to the Euclidean norm of the weight vector. In the past few years a number of approaches and theories have emerged, regarding weight regularization, such as weight normalization \cite{salimans2016weight,huang2017projection} and weight orthogonalization \cite{huang2018orthogonal,bansal2018can,huang2020controllable}. The most popular goal for regularization methods has been weight orthogonalization. The reason is that orthogonal weights have very nice properties in relation to deep network training \cite{bansal2018can,pmlr-v80-chen18i,huang2020controllable}. For example, orthogonal weights can obtain dynamical symmetry \cite{huang2020controllable} which accelerates convergence. Nonetheless, it is not always possible to have orthogonal weights, due to the fact that in many DNN architectures, the weight matrices are over-complete. Moreover, hard-orthogonalization can restrict the learning capacity of a network \cite{bansal2018can}. As an oversimplified, intuitive example of learning capacity restriction we show the schematic visualization of the weights of two $3\times3$, one-channel kernels in Figure \ref{fig:exKernels}. A combination of (a) with a ReLU activation function would result in upper edge detection, while the combination of (b) with a ReLU activation function would result in lower edge detection. The inner product of these filters is -1 and thus wouldn't be possible to have both kernels in a layer restricted to orthogonal weight matrices. Thus, we believe that in some cases weight orthogonalization can be too restrictive, hurting the final performance. In order to overcome this limitation some authors, like \cite{bansal2018can}, reduce the effect of regularization after some point during training, or drop it completely.\par
In this paper we focus on weight regularization by imposing a normalization constraint and propose a new soft-regularization method pushing the weight matrix into the Oblique manifold \cite{absil2006joint,huang2017projection}. To that end we propose a loss function, the norm loss. We evaluate our method with state-of-the-art networks on standard research benchmarks, i.e. CIFAR-10, CIFAR-100 \cite{krizhevsky2009learning} and ImageNet 2012 \cite{russakovsky2015imagenet} and show that networks trained with the norm loss achieve comparable to the state-of-the-art performance and in some cases superior to it, while having negligible computational overhead. To the best of our knowledge we are the first to propose this regularization loss function.\par

The rest of this paper is organized as following. In Section \ref{sec:relatedWork} we discuss the related to our approach research, in Section \ref{sec:preliminaries} we discuss the theory on which our method is based and in Section \ref{sec:proposedMethod} we describe the proposed method. In Section \ref{sec:experiments} we discuss the experimental procedure and show our results and finally in Section \ref{sec:conclusion} we derive our conclusions.\par

\section{Related Work}
\label{sec:relatedWork}

There are many methods that try and alleviate the scaling-based weight space symmetry and the exploding gradients problems. Most of them can be classified into one of two groups, namely hard-regularization and soft-regularization \cite{huang2020controllable}. A very well known approach is the inclusion of the weight decay loss to the total loss which imposes a penalty to the magnitude of the weight vectors \cite{krizhevsky2012imagenet}. It can be considered as a soft-regularization method, since it applies a small change to the weights or the gradients (according to the implementation) in every step. Although this method does address the exploding gradients issue, it does not address the scaling-based weight space symmetry problem. Most methods try to normalize the weights, i.e. force them to have unit norm, make them orthogonal, i.e $\mathbf{W}^T\mathbf{W}  = I*\lambda, $ where $ \lambda$ is a small scaling factor, or both, i.e. $\mathbf{W}^T\mathbf{W} = I$. 
Another work proposed to use the norm of the CNN Jacobian as a training to regularize the models \cite{sokolic2017robust}, while \cite{yoshida2017spectral} proposed spectral norm regularization, which penalizes the high spectral norm of weight matrices.\par
Hard-regularization methods specify hard limits for the weights. For example, \cite{salimans2016weight} divide the weight vector with its norm ($\mathbf{W}^* = \frac{\mathbf{W}}{\|\mathbf{W}\|}$), before applying them to the input, ensuring that the applied weights are normalized. This method is computationally expensive since the normalization happens on the fly. A later work \cite{huang2017projection} ensures that the weights are normalized by normalizing the weight vector every $T$ iterations, i.e. instead of normalizing on the fly they change the original weight vector. This method is more computationally efficient than the previous one and experiments show that it also produces higher performing networks. Moreover, it is the most similar to our approach as we are targeting the same goal but with a soft constraint. Instead of abruptly changing the weights to force unit norm, we add a term to the loss function (instead of the weight decay) to smoothly guide the weights towards unit norm.\par
Other regularization methods try to force the weights to be orthogonal to each other, i.e $\mathbf{W}^T\mathbf{W}=I$ \cite{huang2018orthogonal,huang2020controllable}. Some works \cite{vorontsov2017orthogonality,bansal2018can,xie2017all,balestriero2018spline} apply soft regularization methods by utilizing the standard Frobenius norm and define a term in the training loss function which requires the Gram matrix of the weight matrix to be close to identity, i.e.:
\begin{equation}
L = \lambda\|\mathbf{W}^T\mathbf{W}-I\|
\end{equation}
Although this is an efficient method, when considering over-complete weight vectors it can only be a rough approximation \cite{bansal2018can}. In order to overcome this issue, \cite{bansal2018can} proposed and tested a number of different regularization terms, all focused around weight orthogonalization. Our method is similar to the aforementioned, in the sense that we apply a loose regularization term on the loss function. Our is a bit softer regularization since we only loosely constrain the norm of the weight vectors and do not force them to be orthogonal. Although orthogonal weights have many nice properties, they also restrict the learning capacity of the weights \cite{huang2020controllable}.\par
One of the first works that implemented weight orthogonalization with hard regularization, did so for only fully connected layers \cite{harandi2016generalized}. They defined the generalized back propagation algorithm and compute gradients on the Stiefel manifold. Though this procedure requires the form of Riemannian gradient and a retraction mechanism, which are  computationally intensive. In a later work \cite{ozay2016optimization} extended this approach to convolutional layers as well. \cite{huang2018orthogonal}, \cite{huang2020controllable} propose re-parameterization techniques in order to overcome the limitations of a retraction mechanism. \cite{huang2018orthogonal} uses eigen decomposition to calculate the transformation, whilst \cite{huang2020controllable} used the Newtonian iteration (ONI), which is shown to be more stable and more computationally efficient. The last method also allows for weights to be orthogonal but not normalized, i.e. $\mathbf{W}^T\mathbf{W}  = I*\lambda$. The approach proposed in this paper provides a softer regularization than all aforementioned, with the exception of weight decay, whilst having negligible computational overhead.\par

\section{Preliminaries}
\label{sec:preliminaries}

Given a set of corresponding sets $\{\mathbf{X}, \mathbf{Y}\}$, where every $x_i\in \mathbf{X}$ has a unique corresponding value, or ground-truth label, $y_i\in \mathbf{Y}$, a neural network $f$ should predict the value $y_i$ for the corresponding input $x_i$ with a set of model parameters $\mathbf{W}$. For a given set of model parameters, the neural network output, i.e. the predicted values $\tilde{y_i}=f(x_i,\mathbf{W})$, does not match the desired output $y_i$. If $\mathbf{E}$ is the discrepancy between the $\mathbf{Y}$ and the predicted values $\tilde{\mathbf{Y}}$, the aim of the network training process is to find a set of parameters $\mathbf{W}$ that minimize $E(\mathbf{Y}, \tilde{\mathbf{Y}})$. This is done with the help of a differentiable loss function $L(\mathbf{Y}, \tilde{\mathbf{Y}})$. The training process is carried out by changing by a small factor the set of parameters $\mathbf{W}$ in the direction that minimizes $L(\cdot)$. For an example $x_i, y_i$ or a set of examples $\{\mathbf{X}, \mathbf{Y}\}$ the direction is given by the gradients of $L$, and the weights $\mathbf{W}$ are updated as follows:
\begin{equation}
\label{eq:euclideanGradient}
\mathbf{W}_{new} = \mathbf{W} - \eta\frac{\partial L}{\partial \mathbf{W}}
\end{equation}
where $\eta\ll 1$ denotes the learning rate. For clarity, we make the following remark: Since the Oblique manifold defines matrices with normalized rows, we construct the weight matrix as a row-matrix of the individual weight vectors of each filter, $\mathbf{W} \in \mathbb{R}^{n \times p}$  where $p$ is the dimensionality of each weight vector of a filter, while $n$ is the number of filters. This corresponds to the transpose of the weight matrix typically used in the orthogonality related literature \cite{bansal2018can}.\par
As shown by \cite{huang2017projection}, the Hessian matrix can be ill-conditioned due to the scaling-based weight space symmetry. In an effort to avoid this issue they propose to optimize the network parameters using the Riemannian optimization \cite{absil2009optimization} over the Oblique manifold. The Oblique manifold $\mathcal{BO}(n, p)$ defines a subset of $\mathbb{R}^{n \times p}$, where for every $\mathbf{W} \in \mathcal{BO}(n, p)$:
\begin{equation}
\label{eq:oblique}
ddiag(\mathbf{W}\mathbf{W}^T) = I
\end{equation}
where $ddiag(\cdot)$ is a function that sets all elements of an input array except the diagonal to zero. The above formulation restricts all rows of the matrix $\mathbf{W}$ to have a norm of one. Notice that imposing the requirement of equation \ref{eq:oblique} is less restricting than the usual orthogonalization requirement $\mathbf{W}\mathbf{W}^T = I$, since it only enforces a unit norm but does not enforce the weight vectors to be orthogonal to each other.\par
Given a set of weights $\mathbf{w}$ of a single neuron, i.e., $\mathbf{w} \in \mathbb{R}^{1 \times p}$, where $\mathbf{w}\mathbf{w}^T = 1$, the Riemannian gradients are given by the following equation \cite{huang2017projection}:
\begin{equation}
\label{eq:riemannianDerivative}
\widehat{\frac{\partial L}{\partial \mathbf{W}}} = \frac{\partial L}{\partial \mathbf{W}} - \left(\mathbf{W}^T\frac{\partial L}{\partial \mathbf{W}}\right)\mathbf{w}
\end{equation}
where the norm of the gradients is bound \cite{huang2017projection}:
\begin{equation}
\label{eq:dominantFactor}
\left\|\frac{\partial L}{\partial \mathbf{W}}\right\| \leq \left\|\left(\mathbf{W}^T\frac{\partial L}{\partial \mathbf{W}}\right)\mathbf{w}\right\|
\end{equation}
From equation \ref{eq:dominantFactor} and experimental evidence they conclude that $\frac{\partial L}{\partial \mathbf{W}}$ is the dominant factor of the derivative in equation \ref{eq:riemannianDerivative} and thus propose to apply the Euclidean gradient (equation \ref{eq:euclideanGradient}) and then project the weights back to the Oblique manifold by normalizing them:

\begin{equation}
\label{eq:weightNormalization}
\mathbf{w}^{new} = \frac{\mathbf{w}}{\|\mathbf{w}\|}
\end{equation}

\section{Proposed method}
\label{sec:proposedMethod}

Inspired by the insights of using the Riemannian gradients described in the previous section, we propose to use a similar normalization for neural network training. However, the hard change in the weights after normalization can cause disturbance in the training process since they abruptly shift the weights from the direction of the original gradients. This can be a problem with large learning rates or large projection period $T$, where $T$ is the number of training steps between each projection operation \cite{huang2017projection}. In order to overcome this issue we propose a soft regularization method that instead of abruptly changing the weights, slowly guides them towards the Oblique manifold, i.e., to have unit norm. We implement that by introducing a normalization loss, or norm loss (nl):
\begin{equation}
\label{eq:normLoss}
L_{nl} = \sum_{c_o = 1}^{C_o}\left(1 - \sqrt{\sum_{c_i=1}^{C_i}\sum_{i=1}^{F_h}\sum_{j=1}^{F_w}w_{ijc_ic_o}^2} \right)^2
\end{equation}
where $F_w, F_h, C_i, C_o$ are the filter (or weight vector) width, height, number of input and number of output channels respectively. The loss is penalizing the weight vector of each neuron if its Euclidean norm is different from one. The final loss function then becomes:
\begin{equation}
\label{eq:totalLoss}
L_{total} = L_{target} + \lambda_{nl}\cdot L_{nl}
\end{equation}
where $\lambda_{nl}$ a small factor that determines how strong the regularization will be and $L_{target}$ is the loss function defined by the task to be solved, e.g., triplet loss function, cross entropy loss etc.\par

\subsection{Connection to weight decay}
\label{sec:conWeightDeacay}
The weight decay is similar to our loss since it introduces a small penalty on the magnitude of the weights. The loss function for the weight decay is given by:

\begin{equation}
L_{wd} = \sum_{c_o = 1}^{C_o} \sum_{c_i=1}^{C_i} \sum_{i=1}^{F_h} \sum_{j=1}^{F_w} w_{ijc_ic_o}^2
\end{equation}

There are two main differences. Firstly, weight decay penalizes the absolute magnitude of the weight vector while norm loss penalizes the deviation to having unit  norm. This means that in the case the norm is smaller than one, our method will try to increase it, whilst the weight decay will continue pushing to decrease it. This makes a difference in situations where some components are rarely being utilized (e.g., due to nonlinearities such as the ReLU) in which case the main loss changing these components is the weight decay loss, resulting in very small weights that might never recover. The second difference is that it applies to all components of a layer uniformly, whilst the norm loss differentiates between the vectors of each output channel. This can be seen by the derivatives of each method. The derivatives for each component of the weight matrix are given by:
\begin{equation}
\label{eq:weightDecayDer}
\frac{\partial L_{wd}}{\partial w_{ijc_ic_o}} = 2w_{ijc_ic_o}
\end{equation}

From Equation \ref{eq:normLoss} it is easy to derive the gradients of the norm loss (Derivation can be found in the appendix):
\begin{equation}
\label{eq:normLossDer}
\frac{\partial L_{nl}}{\partial w_{ijc_ic_o}} = 2w_{ijc_ic_o}\left (1 - \frac{1}{\|w_{c_o}\|}\right)
\end{equation}
where $\|w_{c_o}\|$ is the norm of the weight vector of a single neuron, with index $c_o$. Comparing equations \ref{eq:weightDecayDer} and \ref{eq:normLossDer} we can see that effectively norm loss can be seen as an extension of the weight decay where the weight decay factor and its sign are regulated during training by the norm of the weight vector. This is explicitly visible from the overall update rule (combining equations \ref{eq:euclideanGradient}, \ref{eq:totalLoss}, \ref{eq:normLossDer}):
\begin{equation}
\label{eq:upadateRule}
\mathbf{w}^{new} = \mathbf{w} -\eta\lambda_{nl}2\left (1 - \frac{1}{\|w_{c_o}\|}\right)\mathbf{w} - \eta\frac{\partial L_{target}}{\partial \mathbf{w}}
\end{equation}

\subsection{Computational cost}

For a convolutional layer with $C_o$ filters of shape $F_h\times F_w\times C_i$, the computational cost of weight decay is two operations per weight, thus $2\cdot C_o\cdot F_h\cdot F_w\cdot C_i$. For the norm loss it is $3 \cdot C_o \cdot (F_h\cdot F_w \cdot C_i) + C_o + 2\cdot C_o\cdot F_h\cdot F_w\cdot C_i$. The overhead then is $3 \cdot C_o \cdot (F_h\cdot F_w \cdot C_i) + C_o$. The computational cost of a convolutional layer is $6 \cdot m \cdot C_o \cdot C_i \cdot F_h \cdot F_w \cdot I_h \cdot I_w$, where $m, I_h, I_w$ are the number of images in a mini batch, and the input (to the layer) image height and width respectively. The computational overhead of the norm loss is orders of magnitude smaller than the computational cost of a convolutional layer with weight decay, for usual network and image input sizes.

\section{Experiments}
\label{sec:experiments}

We evaluate our method on three well known benchmarks, i.e., CIFAR-10, CIFAR-100 and ImageNet2012. CIFAR-10 consists of 50K training and 10K test $32\times 32$ natural images, divided into 10 classes. CIFAR-100 also consists of 50K training and 10K test natural images with the same resolution, but in this dataset they are divided into 100 classes. For both CIFAR-10 and CIFAR-100 we evaluate using classification error on the designated test sets. The ImageNet 2012 is a large scale image recognition dataset. The train set consists of 1.281 million natural images of arbitrary resolution and aspect ratio. The images are divided into 1000 classes. The validation set consists of 50K images, i.e. 50 images per class. We evaluate our method on the validation set top 1 and top 5 error rates, as is common practice in the field.\par
We utilize two different state-of-the-art architectures, namely the ResNet \cite{he2016deep} and the wide ResNet (WRN) \cite{BMVC2016_87}, since they are usual test cases for weight regularization methods \cite{huang2017projection,huang2018orthogonal,huang2020controllable,bansal2018can}. Both ResNet and WRN have been defined for many different sizes with different learning capacity. Unfortunately, training such big networks is very computationally expensive and thus we choose only one architecture per model. For all our three benchmarks we utilize the cross-entropy loss function as our target loss function ($L_{target}$).\par
We compare our method with state-of-the-art approaches, like weight decay (wd), weight normalization (WN)\cite{salimans2016weight}, projection based weight normalization (PBWN)\cite{huang2017projection}, orthogonalization with Newton iteration (ONI)\cite{huang2020controllable}, orthogonal linear module (OLM)\cite{huang2018orthogonal}.\par
\subsection{Regularization factor}
\begin{figure}[h]
\centering
\includegraphics[width=\linewidth]{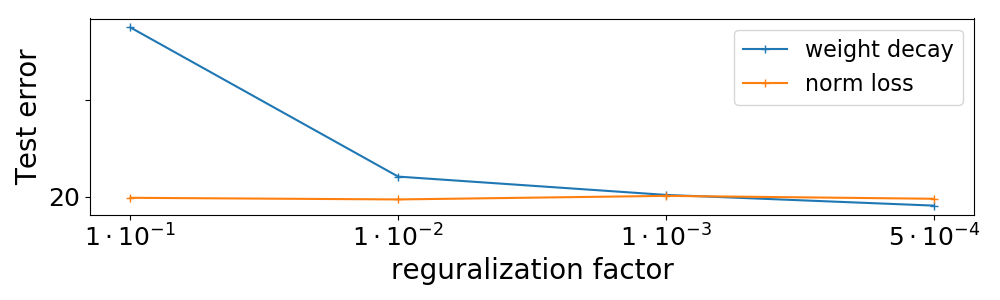}
\caption{Test error of WRN-28-10 on CIFAR-100 with weight decay and norm loss for different regularization factors.}
\label{fig:reg_factor}
\end{figure}

With our first experiment we evaluate the effect of the regularization factor $\lambda_{nl}$ on the training. As a benchmark we use the CIFAR-100 dataset. We train for four different values of $\lambda_{nl}$ and plot our results in Figure \ref{fig:reg_factor}. It is apparent that the effect of $\lambda_{nl}$ on the training is much smaller than that of $\lambda_{wd}$ in the case of weight decay. We believe that this results from the regularization of the $\lambda_{nl}$ factor discussed before (equation \ref{eq:upadateRule}).\par
\begin{figure}[h]
\centering
\includegraphics[width=\linewidth]{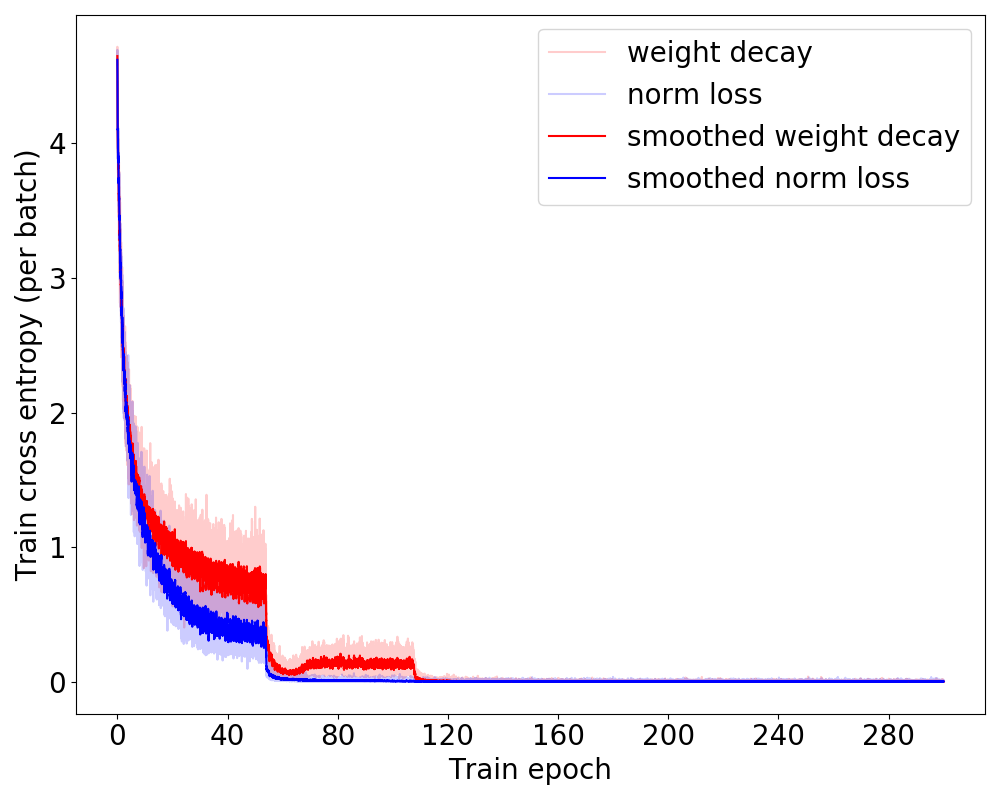}
\caption{Evolution of cross entropy during training on CIFAR-100 for weight decay and norm loss. The regularization factor for both runs is 5e-4. The shaded lines are the true lines, whilst the non-shaded are smoothed version of the original (averaged over 19 steps) for more comprehensive visualization.}
\label{fig:trainCrossEntropy_cifar100}
\end{figure}
Figure \ref{fig:trainCrossEntropy_cifar100} shows the evolution of the train cross entropy (per batch) during training, for the case of $\lambda=5\cdot 10^{-4}$. We can see that with the norm loss, the networks are being trained faster than with weight decay, even in the case where the weight decay training has marginally higher accuracy in the end of training (see Figure \ref{fig:reg_factor}).\par
\subsection{Batch size}
\begin{figure}[h]
\centering
\includegraphics[width=\linewidth]{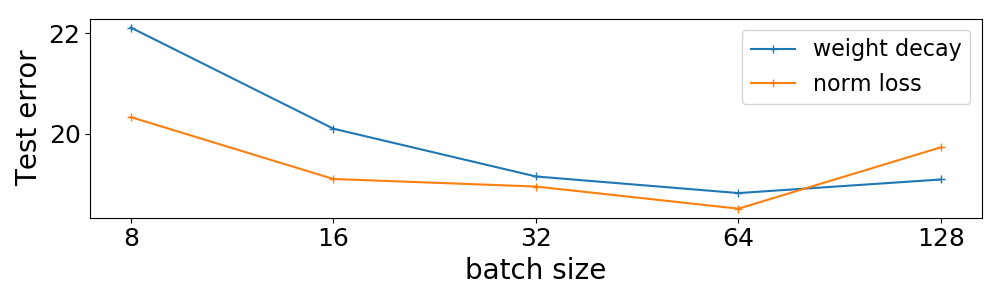}
\caption{Test error of WRN-28-10 on CIFAR-100 with weight decay and norm loss for different batch sizes.}
\label{fig:batch_size}
\end{figure}
The next experiment is to test the training behavior for different batch sizes. We train networks with both weight decay and our method for batch sizes $\{8,16,32,64,128\}$ on CIFAR-100. The performance of the trained networks on the test set can be seen in Figure \ref{fig:batch_size}. We can see that the norm loss has more steady behavior than the weight decay. We can still see some dependence on the batch size, which is expected since our networks utilize batch normalization. Moreover, although for most batch sizes tested the networks trained with norm loss show better performance, only for batch size 128 the opposite is true. Finally, we can see that the highest overall performance for both methods is achieved by the networks trained with batch size 64.\par
\subsection{CIFAR-10}
On this dataset we train on the ResNet110 and WRN-28-10. As is common practice \cite{he2016deep,BMVC2016_87} we train the networks using SGD with Momentum of 0.9 and a batch size of 128. The initial learning rate is set to 0.1. We follow a learning schedule close to the one used in \cite{BMVC2016_87}. When training the WRN, we drop the learning rate by a factor of 5 at epochs 53, 107, 230 and we train for a total of 300 epochs. For the ResNet we train for 164 epochs and reduce the learning rate at epochs 82 and 123. We follow the standard preprocessing for training as in \cite{he2016deep,BMVC2016_87,bansal2018can}, i.e. padding each train example by four pixels and getting a random $32\times32$ crop. Both train and test sets are mean and std normalized \cite{BMVC2016_87}. We train five times and report the mean as well as the best run. For both networks, the regularization factor $\lambda_{nl}$ is set to 0.01 while the weight decay factor $\lambda_{wd}$ is set to $10^{-4}$ for the ResNet and $5\cdot 10^{-4}$ for the WRN, as in the original papers \cite{he2016deep,BMVC2016_87}. The results can be seen in Table \ref{tab:cifar10_results}.\par
\begin{table}[h]
\renewcommand{\arraystretch}{1.3}
\caption{Performance of different methods on CIFAR-10 test set for the ResNet110 and WRN-28-10. In parentheses are the mean or median over some runs given by the authors of the respective papers. Outside parentheses the best accuracy (if shown by the authors). For methods denoted by *, the performance is given in the respective papers.}
\label{tab:cifar10_results}
\centering
\begin{tabular}{lrr}
\toprule
model & reg. method & error \\
\midrule
 ResNet110 & wd & 6.32 (6.568)\\
 ResNet110 \cite{he2016deep}* & wd & 6.43 (6.61) \\
 ResNet110 \cite{salimans2016weight}* & WN & - (7.56) \\ 
 ResNet110 \cite{huang2017projection}* & PBWN & - (6.27) \\ 
 ResNet110 \cite{huang2020controllable}* & ONI & - (6.56) \\ 
 ResNet110 (Ours) & nl & 5.9 (5.996) \\ 
\midrule
 WRN-28-10 & wd & 3.9 (3.966) \\
 WRN-28-10 \cite{BMVC2016_87}* & wd & - (3.89) \\
 WRN-28-10 \cite{huang2018orthogonal}* & OLM\ & - (3.73) \\
 WRN-28-10 \cite{huang2018orthogonal}* & OLM-L1\ & - (3.82) \\
 WRN-28-10 (Ours) & nl & 4.47 (4.662) \\ 
 \bottomrule
\end{tabular}
\end{table}

When training the ResNet110 we manage to outperform all other regularization methods we are aware of, that tested on the same network \cite{huang2020controllable,salimans2016weight,huang2017projection}. We also outperform the reported accuracy of the SRIP regularization method of \cite{bansal2018can}. Due to large difference with our and their baseline performance, we do not consider it a valid comparison and thus their result is omitted from Table \ref{tab:cifar10_results}.\par
\begin{figure}[h]
\centering
\includegraphics[width=\linewidth]{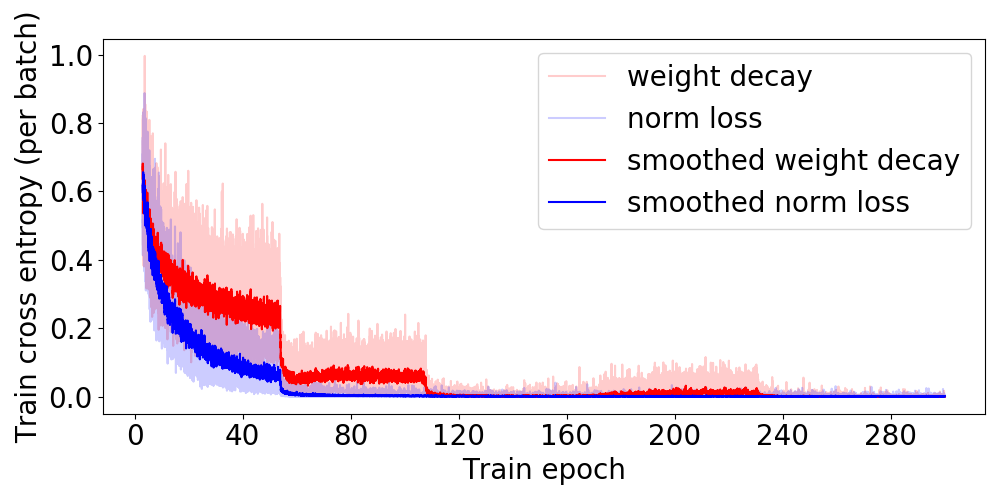}
\caption{Evolution of cross entropy during training on CIFAR-10 for weight decay and norm loss for the WRN-28-10. The shaded lines are the true lines, whilst the non-shaded are smoothed version of the original (averaged over 19 steps) for more comprehensive visualization.}
\label{fig:trainCrossEntropy_cifar10}
\end{figure}
\begin{figure}[h]
\centering
\includegraphics[width=\linewidth]{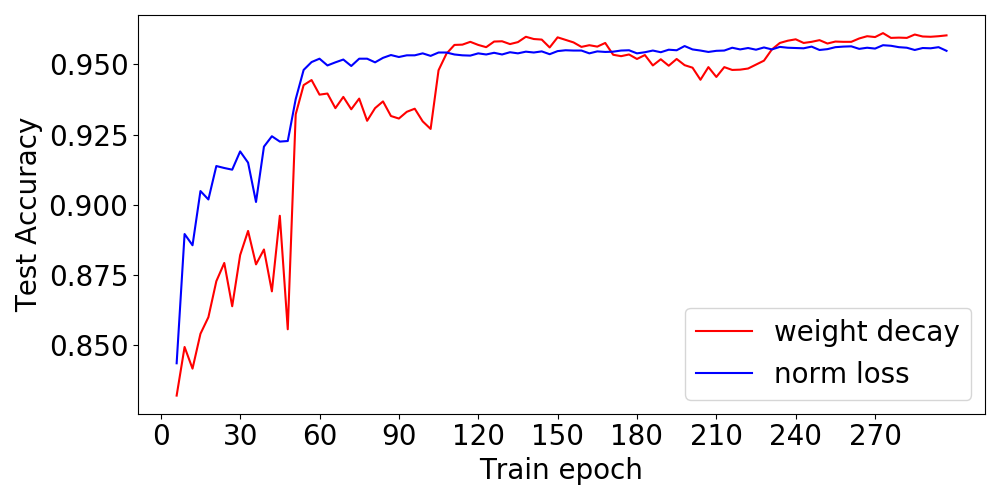}
\caption{Evolution of test accuracy during training on CIFAR-10 for weight decay and norm loss for the WRN-28-10.}
\label{fig:testAccuracy_cifar10}
\end{figure}

When training the WRN-28-10 we observe a big decrease in performance over our benchmark. Figures \ref{fig:trainCrossEntropy_cifar10}, \ref{fig:testAccuracy_cifar10} show the training process of the WRN-28-10 on the CIFAR-10 with weight decay and norm loss. We can see that even in this case, where the final performance of weight decay is better, the network that utilizes the norm loss is converging much faster and shows more stable behavior. More experimentation is needed to understand why in this specific scenario the final performance is worse than the baseline. For example, a schema like the one used in \cite{bansal2018can} could be used, where from a certain epoch on the regularization is minimized or even dropped completely.\par


In order to evaluate the computational overhead of our method, we report that training the ResNet110 on CIFAR-10 with weight decay takes, on average, 4.56 hours while training the same network with norm loss takes, on average, 4.79 hours. All aforementioned experiments ran on an Intel(R) Xeon(R) CPU E5-2699 v4 @ 2.20GHz and an NVIDIA GTX 1080Ti graphics card.\par

\subsection{CIFAR-100}

As with CIFAR-10, for this section we use the same hyper parameters as in \cite{BMVC2016_87}, with a slightly different learning rate schedule. We train the WRN using a batch size of 64 instead of 128, because it results in better performance even for the baseline method. The regularization factor $\lambda_{nl}$ was set to $10^{-3}$ and the $\lambda_{wd}$ to $5\cdot 10^{-4}$. We train for a total of 448 epochs and reduce the learning rate by a factor of 5 in epochs 77, 153 and 307. We train the ResNet for 300 epochs and drop the learning rate at epochs 53, 107, 230. We train 5 times and report the accuracy of the mean and best run. The results are shown in Table \ref{tab:cifar100_results}. We also trained with the learning rate scheduler used for ResNet110 on CIFAR-10, i.e. 164 epochs and reduce the learning rate at epochs 82 and 123, but the aforementioned scheduler produced results for the baseline that match the literature and thus is reported on Table \ref{tab:cifar100_results}. For clarity we also report the results using the CIFAR-10 scheduler: Average error for the weight decay is $28.094$ with a best run of $27.56$ whilst for the norm loss the average error is $25.878$ with a best run of $25.2$. For all ResNet experiments, the regularization factor $\lambda_{nl}$ was set to $10^{-2}$ and the $\lambda_{wd}$ to $10^{-4}$.\par
The norm loss manages to produce better performance than most methods for both ResNet110 and WRN-28-10. Although in the case of CIFAR-10 we managed to outperform the method developed in \cite{bansal2018can} (results were omitted from the tables due to large difference of their and our baseline performance), in the case of CIFAR-100 their methods outperform our own. For the same reason as before their results are omitted (25.42\% vs 27.49\% accuracy for the baseline ResNet110). It should be noted that they used a different optimizer for their experiments, the Adam optimizer \cite{DBLP:journals/corr/KingmaB14}.

 

 

\begin{table}[h]
\renewcommand{\arraystretch}{1.3}
\caption{Performance of different methods on CIFAR-100 test set for the ResNet110 and WRN-28-10. In parentheses are the mean or median over some runs given by the authors of the respective papers. Outside parentheses the best accuracy (if shown by the authors). For methods denoted by *, the performance is given in the respective papers. }
\label{tab:cifar100_results}
\centering
\begin{tabular}{lrr}
\toprule
model & reg. method & error \\
\midrule
 ResNet110 & wd & 27.9 (28.398) \\
 ResNet110 \cite{salimans2016weight}* & WN & - (28.38) \\ 
 ResNet110 \cite{huang2017projection}* & PBWN & - (27.03) \\ 
 ResNet110 (Ours) & nl & 26.24 (26.526) \\ 
\midrule
 WRN-28-10 & wd & 18.85 (19.138) \\
 WRN-28-10 \cite{BMVC2016_87}* & wd & - (18.85) \\
 WRN-28-10 \cite{huang2018orthogonal}* & OLM & - (18.76) \\
 WRN-28-10 \cite{huang2018orthogonal}* & OLM-L1 & - (18.61) \\
 WRN-28-10 (Ours) & nl & 18.57 (18.648) \\ 
 \bottomrule
\end{tabular}
\end{table}

\subsection{ImageNet}

For the ImageNet dataset, we utilize the ResNet50 architecture to evaluate our method. We use the same data augmentation and hyper parameters (for batch normalization, dropout rates, etc.) as the implementation of \cite{huang2017projection} on GitHub\footnote{https://github.com/huangleiBuaa/NormProjection}. We train using SGD with Nesterov momentum of 0.9. The learning rate is initialized at 0.1 and divided by 10 every 30 epochs. We train with a batch size of 128, whilst due to GPU memory limitations, the batch normalization parameters are trained on half, i.e. 64 examples. The regularization factor for both methods, i.e. $\lambda_{nl}, \lambda_{wd}$ is set to $10^{-4}$ The top-1 and top-5 error rates \footnote{Top-1 error is 100 - classification accuracy. Top-5 error counts a correct classification if the ground truth label is in the top-5 predictions of the network.} of the networks on the ImageNet 2012 validation set are shown in Table \ref{tab:imageNet_results}.\par
The results are shown in Table \ref{tab:imageNet_results}. The first two rows show the  performance of the baseline method as reported in literature\cite{huang2020controllable} and our attempt at reproducing it. As it can be seen, there is a large difference between the results as our error is much larger than the one reported in literature. The origin of this is currently not clear to us and therefore a direct comparison of performance numbers is not warranted. Nonetheless, we can observe that for our implementations the norm loss approach improves both TOP-1 and TOP-5 performances substantially over weight decay.\par

\begin{table}[t!]
\renewcommand{\arraystretch}{1.3}
\caption{Top-1 and Top-5 error rates of different methods on ImageNet validation set for the ResNet50. For methods denoted by *, the performance is given in the respective papers.}
\label{tab:imageNet_results}
\centering
\begin{tabular}{lrrr}
\toprule
model & reg. method & Top-1 error & Top-5 error\\
\midrule
 ResNet50 & wd & 25.29 & 7.86 \\
 ResNet50 \cite{huang2020controllable}* & wd & 23.85 & - \\ 
 ResNet50 \cite{huang2020controllable}* & ONI & 23.30 & - \\ 
 ResNet50 (Ours) & nl & 24.34 & 7.44 \\ 
 \bottomrule
\end{tabular}
\end{table}

\section{Conclusion}
\label{sec:conclusion}
In this work we proposed a new soft-regularization method, that tries to guide the weight vector towards the Oblique manifold. It accomplishes that by utilizing the proposed norm-loss function as regularization during the neural network training.\par
We evaluate our method on standard benchmarks, i.e. CIFAR-10, CIFAR-100 and ImageNet 2012 and compare the performance to the state-of-the-art regularization methods. Our method accelerates the convergence speed of networks and leads to a more robust, i.e.\ less sensitive training process with regard to the hyperparameters settings for the regularization factor and the batch size. While having a negligible computational overhead over weight decay during training, and no extra computational cost during inference, our method has comparable performance to the state-of-the-art and some cases even higher, while only for one setup our method is under performing, i.e. WRN-28-10 on CIFAR-10. \par

\section*{Acknowledgment}

This work is part of the research program DAMIOSO with project number 628.006.002, which is partly financed by the Netherlands Organization for Scientific Research (NWO) and partly by Honda Research Institute-Europe (GmbH).



%
\bibliographystyle{IEEEtran}
\bibliography{normLoss}

\newpage
\onecolumn
\section*{Appendix}

Derivation of norm loss gradients:
\begin{equation}
\begin{split}
&L_{nl} = \sum_{c_o = 1}^{C_o}\left(1 - \sqrt{\sum_{c_i=1}^{C_i}\sum_{i=1}^{F_h}\sum_{j=1}^{F_w}w_{ijc_ic_o}^2} \right)^2 \Rightarrow \frac{\partial L_{nl}}{\partial {w_{ijc_ic_o}}} = \frac{\partial}{\partial {w_{ijc_ic_o}}} \left( \sum_{c_o = 1}^{C_o}\left(1 - \sqrt{\sum_{c_i=1}^{C_i}\sum_{i=1}^{F_h}\sum_{j=1}^{F_w}w_{ijc_ic_o}^2} \right)^2 \right) = \\
& \frac{\partial}{\partial {w_{ijc_ic_o}}}\left(1 - \sqrt{\sum_{c_i=1}^{C_i}\sum_{i=1}^{F_h}\sum_{j=1}^{F_w}w_{ijc_ic_o}^2} \right)^2 = 
2\left(1 - \sqrt{\sum_{c_i=1}^{C_i}\sum_{i=1}^{F_h}\sum_{j=1}^{F_w}w_{ijc_ic_o}^2} \right) \frac{\partial}{\partial {w_{ijc_ic_o}}}\left(1 - \sqrt{\sum_{c_i=1}^{C_i}\sum_{i=1}^{F_h}\sum_{j=1}^{F_w}w_{ijc_ic_o}^2} \right)
\end{split}
\end{equation}

The norm of a weight matrix of a kernel with index $c_o$ is:
\begin{equation}
\|w_{c_o}\| = \sqrt{\sum_{c_i=1}^{C_i}\sum_{i=1}^{F_h}\sum_{j=1}^{F_w}w_{ijc_ic_o}^2}
\end{equation}

From equations (1), (2):

\begin{equation}
\begin{split}
&\frac{\partial L_{nl}}{\partial {w_{ijc_ic_o}}} = 2\left(1- \|w_{c_o}\| \right) \left(-\frac{1}{2\sqrt{\sum_{c_i=1}^{C_i}\sum_{i=1}^{F_h}\sum_{j=1}^{F_w}w_{ijc_ic_o}^2}}  \right)\frac{\partial}{\partial {w_{ijc_ic_o}}}\left( \sum_{c_i=1}^{C_i}\sum_{i=1}^{F_h}\sum_{j=1}^{F_w}w_{ijc_ic_o}^2 \right) \Rightarrow\\
&\frac{\partial L_{nl}}{\partial {w_{ijc_ic_o}}} = 2\left(1- \|w_{c_o}\| \right) \left(-\frac{1}{2\|w_{c_o}\|}  \right)2w_{ijc_ic_o} = 2w_{ijc_ic_o}\left(1 - \frac{1}{\|w_{c_o}\|} \right)
\end{split}
\end{equation}

\end{document}